# Deep Learning-Accelerated 3D Carbon Storage Reservoir Pressure Forecasting Based on Data Assimilation Using Surface Displacement from InSAR

Hewei Tang[1*], Pengcheng Fu[1], Honggeun Jo[1], Su Jiang[1*], Christopher S. Sherman[1], François Hamon[2], Nicholas A. Azzolina[3], and Joseph P. Morris[1]

[1]Atmospheric, Earth, and Energy Division, Lawrence Livermore National Laboratory, Livermore, CA 94550, USA
[2]TotalEnergies E&P Research and Technology, Houston, TX 77002, USA
[3]University of North Dakota, Energy & Environmental Research Center, Grand Forks, ND 58202-9018, USA

* Corresponding authors: tang39@llnl.gov, jiang14@llnl.gov

Key Points:

- Proposed a 3D data assimilation workflow for reservoir pressure forecasting in geologic carbon storage.
- Applied deep learning techniques to significantly improve the efficiency of the workflow.
- Demonstrated the efficacy of using InSAR data as monitoring data to infer reservoir pressure build up.

Abstract

Fast forecasting of the reservoir pressure distribution during geologic carbon storage (GCS) by assimilating monitoring data is a challenging problem. Due to high drilling cost, GCS projects usually have spatially sparse measurements from few wells, leading to high uncertainties in reservoir pressure prediction. To address this challenge, we use low-cost Interferometric Synthetic-Aperture Radar (InSAR) data as monitoring data to infer reservoir pressure build up. We develop a deep learning-accelerated workflow to assimilate surface displacement maps interpreted from InSAR and to forecast dynamic reservoir pressure. Employing an Ensemble Smoother Multiple Data Assimilation (ES-MDA) framework, the workflow updates three-dimensional (3D) geologic properties and predicts reservoir pressure with quantified uncertainties. We use a synthetic commercial-scale GCS model with bimodally distributed permeability and porosity to demonstrate the efficacy of the workflow. A two-step CNN-PCA approach is employed to parameterize the bimodal fields. The computational efficiency of the workflow is boosted by two residual U-Net based surrogate models for surface displacement and reservoir pressure predictions, respectively. The workflow can complete data assimilation and reservoir pressure forecasting in half an hour on a personal computer.

Keywords: data assimilation, deep learning, geologic carbon storage, surrogate modeling, ES-MDA

1. Introduction

Rapid data assimilation and associated reservoir performance forecasting is essential for robust management of many subsurface applications such as geologic carbon storage (GCS), geothermal energy recovery, oil/gas production and ground water resource management (Jiang & Durlofsky, 2020; Jo et al., 2021; Sun et al., 2021; Wu et al., 2021). The process of data assimilation (also known as stochastic inverse modeling) is to estimate the probability distributions of certain unknown parameters for a system by integrating observation data. Forecasting the behavior of the system with quantified uncertainties is the chief goal of such data assimilation. Due to ubiquitous heterogeneity in subsurface porous media, we are usually concerned with a high-dimensional inverse problem that requires a large number of forward model runs to sufficiently capture the uncertainty. In the traditional workflow, these forward model runs and subsequent data assimilation are computationally expensive and time-intensive, which are major impediments to providing rapid data interpretations that can inform decision-making. Accelerating the data assimilation and the forecasting process will enable rapid decision-making regarding operational strategies and result in significant cost savings and more optimal system performance.

Estimating the pressure distribution in GCS reservoirs is important yet challenging. It is essential in the permitting phase as well as the monitoring, reporting, and verification (MRV) plan required under subpart of the US EPA Greenhouse Gas Reporting Program (Burton-Kelly et al., 2021; EPA, 2022). However, the spatially sparse observation data from a limited number of wells can lead to high uncertainties in the prediction of the reservoir pressure's spatial distribution. Accurate prediction of reservoir pressure distribution during $CO_2$ injection requires high-dimensional observation data to track the reservoir pressure changes. Interferometric Synthetic-Aperture Radar (InSAR) has proven to be an effective monitoring technique to track the changes of reservoir surface deformation and thereby to infer reservoir pressure changes. InSAR analysis compares a series of radar images of the same location, collected from orbiting satellites, at different times. Elevation changes of ground surface can be calculated from a detectable change in the phase of the reflected signal (Becker, 2006). InSAR has been widely used to measure ground surface displacements (Alghamdi et al., 2020; Hoffmann et al., 2001, 2003). For GCS applications, InSAR has been successfully applied to measure surface uplift in the In Salah $CO_2$ storage project, located in the central region of Algeria (White et al., 2014). The data indicated an order of 5 mm per year ground surface uplift above the $CO_2$ injection reservoir and an uplift pattern extending several kilometers around the injection wells (Vasco et al., 2008). The uplift magnitude can be explained by pressure-induced, poroelastic expansion of the storage reservoir based on coupled reservoir-geomechanics modeling (Rutqvist et al., 2010). Those studies revealed that InSAR-measured deformation is highly correlated to reservoir pressure changes, a critical fact forming the basis of the current work. Compared with traditional well-based pressure monitoring, InSAR provides high-dimensional observation data at a low cost, potentially reducing the uncertainties in predicting the spatial distribution of reservoir pressure.

In this study, we proposed to develop a computationally efficient workflow to assimilate surface displacement data interpreted from InSAR to calibrate the underlying reservoir models and to predict dynamic reservoir pressure distributions. To obtain a desirable level of fidelity of reservoir pressure, the workflow is designed to calibrate heterogeneous rock properties such as rock facies, porosity, and permeability in three dimensions (3D). Most studies on data assimilation of surface displacement maps have been in two dimensions (2D) due to the high computational cost of 3D modeling. Iglesias & McLaughlin (2012) presented an inverse modeling approach to estimate heterogeneous permeability and elastic moduli in a 2D $40 \times 40$ gridded domain from measurements of surface displacement and production data. Recently,

Alghamdi et al. (2020) developed a high-dimensional Bayesian inversion framework to infer permeability on a 16,896-cell 2D grid from surface deformation measurement. A full 3D data assimilation workflow requires a parameter space of at least $10^5$ degrees of freedom. Such a high-dimensional parameter space renders traditional Bayesian approaches such as rejection sampling and Markov Chain Monte Carlo inapplicable. Ensemble-based methods such as ensemble smoother (ES) and ensemble Kalman filter (EnKF) are efficient alternative methods for high-dimensional inversion. Ensemble-based methods apply first- and second- order statistical information for error propagation, thus avoiding the high computational cost associated with gradient calculation (Ma et al., 2019; Tang, et al., 2021). In this study, we develop the workflow based on an iterative ES method called Ensemble Smoother with Multiple Data Assimilation (ES-MDA), developed by Emerick & Reynolds (2013). The ES-MDA approach has been demonstrated to have satisfactory performance with low computational cost (Chen et al., 2020; Emerick & Reynolds, 2013; Liu et al., 2020).

To develop a computationally efficient workflow for rapid reservoir pressure forecasting, two major challenges must be addressed. The first challenge is to parameterize a high-dimensional 3D parameter space which includes facies, porosity, and permeability distributions. As explained in Section 4.1, we are concerned with bimodally distributed non-Gaussian fields represented by heterogeneous porosity/permeability distributions in each facies. Although deep neural networks such as generative adversarial networks (GAN) and variational autoencoders (VAE) have been recently applied in non-Gaussian field parameterization, studies that parameterize a 3D system with heterogeneous properties within individual facies are still scarce. Two related works emerged recently and are both based on the successful application of convolutional neural networks. The first work applied a convolutional adversarial autoencoder network to generate an unconditional 3D non-Gaussian conductivity field (Mo et al., 2020). The second work is a variant of the principal component analysis (PCA) method, in which a convolutional neural network called "C3D net" is applied as a post-processor to PCA to improve the parameterization performance for a wide range of geological systems (Liu & Durlofsky, 2021). We adopted the parameterization method proposed in Liu & Durlofsky (2021), which has proven efficient and able to sufficiently honor the flow behavior of the system.

The second challenge is to develop a computationally efficient surrogate model to predict surface displacement (a 2D map) from 3D non-Gaussian parameter distributions. The high-dimensional and non-Gaussian nature of the problem makes traditional surrogate models such as Gaussian process inapplicable. Deep neural networks have been demonstrated as an effective solution for high-dimensional surrogate modeling by treating surrogate modeling as an image-to-image regression task. Mo et al. (2019) developed a deep dense convolutional neural network to predict $CO_2$ plume migration in 2D Gaussian permeability fields. Identifying the large errors resulting when applying the model to 3D non-Gaussian conductivity fields, Mo et al. (2020) improved the model by adding multilevel residual blocks. Tang et al. (2021b) developed a 3D recurrent residual U-Net (R-U-Net) model to predict dynamic saturation and pressure fields for a binary channel system. The authors later applied the same neural network to predict dynamic saturation, pressure, and displacement fields in GCS applications with Gaussian permeability distributions (Tang et al., 2021a). In the current work, we develop a new R-U-Net architecture to predict 2D surface displacement maps directly from 3D non-Gaussian permeability and porosity distributions.

In summary, the major innovative contributions of this study include: 1) we develop a data assimilation workflow to predict high-resolution dynamic pressure distributions in GCS reservoirs using high-dimensional InSAR-based surface displacement data rather than sparse pressure measurements from wells; 2) we perform computationally efficient inversion for 3D non-Gaussian parameter fields; and 3) we develop a 3D-to-2D R-U-Net structure to directly predict surface displacement in the form of 2D maps.

The rest of the paper is organized as follows. First, we introduce the data assimilation and forecasting workflow with its major components. The forward numerical model and the relevant surrogate models applied in the workflow are introduced subsequently. We employ a commercial-scale 3D synthetic case to demonstrate the implementation of the workflow. Numerical results for this case are presented and discussed. In the last section, we summarize our conclusions

## 2. Data Assimilation Workflow

Fig.1 presents an overview of the data assimilation and forecasting workflow. There are four major components in the workflow: 1) model parameterization using the CNN-PCA approach, 2) data space dimension reduction using PCA, 3) ES-MDA inversion for model parameters based on InSAR data, and 4) surface displacement and reservoir pressure predictions with deep-learning surrogate models. We will describe the details of the first three components in the following sub-sections. The numerical forward model and the relevant deep learning surrogate models will be introduced in Section 3.

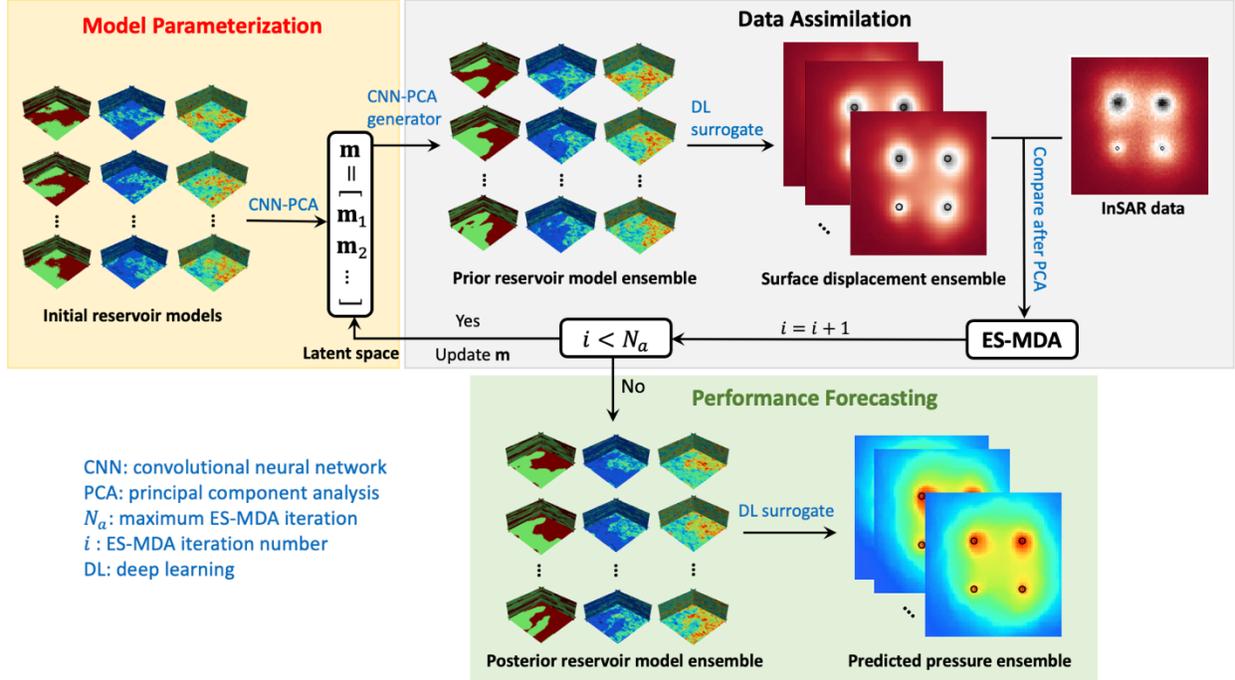

Fig.1. Overview of the workflow for InSAR data assimilation and reservoir pressure forecasting. Each realization in the reservoir model ensemble is represented by a facies field, a porosity field, and a permeability field.

2.1 Parameterization of initial reservoir models

The purpose of model parameterization is to create a low-dimensional representation for initial reservoir models including facies, porosity, and logarithmic permeability distributions. To construct a traditional PCA representation, an ensemble of $N_r$ realizations of vectorized geological variables $\mathbf{m} \in \mathbb{R}^{N_c}$ ($N_c$ is the number of grid cells) is generated. These vectors are assembled into a data matrix $\mathbf{X} \in \mathbb{R}^{N_c \times N_r}$

$$\mathbf{X} = \frac{1}{\sqrt{N_r - 1}} [\mathbf{m}_1 - \boldsymbol{\mu}_\mathbf{m}, \ \mathbf{m}_2 - \boldsymbol{\mu}_\mathbf{m}, \ \ldots, \ \mathbf{m}_{N_r} - \boldsymbol{\mu}_\mathbf{m}], \tag{1}$$

Where $\boldsymbol{\mu}_\mathbf{m}$ is the mean of the $N_r$ realizations. A singular value decomposition yields $\mathbf{X} = \mathbf{U}\boldsymbol{\Sigma}\mathbf{V}^\mathbf{T}$, where $\mathbf{U} \in \mathbb{R}^{N_c \times N_r}$ and $\mathbf{V} \in \mathbb{R}^{N_r \times N_r}$ are the left and right singular matrices, and $\boldsymbol{\Sigma} \in \mathbb{R}^{N_r \times N_r}$ is a diagonal matrix containing singular values ($\Sigma_{ii}$) of $\boldsymbol{\Sigma}$. We select $n$ largest components in $\boldsymbol{\Sigma}$ and the corresponding columns in $\mathbf{U}$, denoted as $\boldsymbol{\Sigma}_n$ and $\mathbf{U}_n$, respectively. A new PCA realization with features similar to that of original $N_r$ realizations can be generated from any given low-dimensional vector $\boldsymbol{\lambda} \in \mathbb{R}^{n \times 1}$ sampled from $N(0, \mathbf{I}_n)$:

$$\mathbf{m}_{\text{PCA}} = \boldsymbol{\mu}_\mathbf{m} + \mathbf{U}_n \boldsymbol{\Sigma}_n \boldsymbol{\lambda} \tag{2}$$

$\boldsymbol{\lambda}$ is also known as latent space vector. We can reconstruct the original realizations through the projected latent space vector ($\tilde{\boldsymbol{\lambda}}_i$):

$$\tilde{\boldsymbol{\lambda}}_i = \boldsymbol{\Sigma}_n^{-1} \mathbf{U}_n^T (\mathbf{m}_i - \boldsymbol{\mu}_\mathbf{m}), \quad i = 1, \ldots, N_r \tag{3}$$

$$\tilde{\mathbf{m}}_{\text{PCA},i} = \boldsymbol{\mu}_\mathbf{m} + \mathbf{U}_n \boldsymbol{\Sigma}_n \tilde{\boldsymbol{\lambda}}_i, \quad i = 1, \ldots, N_r \tag{4}$$

The choice of the latent space vector dimension ($n$) can be informed by the criterion $\sum_{i=1}^{n} \Sigma_{ii} / \sum_{i=1}^{N_r} \Sigma_{ii} \geq T$, where $T$ represents the percentage of the total variance being preserved in the original data (Vo & Durlofsky, 2014). If $n = N_r$, $\tilde{\mathbf{m}}_{\text{PCA},i}$ will exactly match $\mathbf{m}_i$. This traditional PCA workflow is directly applicable to geological variables following a multi-Gaussian distribution. For non-Gaussian distributions (such as facies), the traditional PCA cannot be directly applied.

In this work, we will perform parameterization in two steps: the first step is to parameterize facies distributions using a CNN-PCA approach, and the second step is to parameterize porosity and permeability distributions using the traditional PCA approach. The CNN-PCA approach applies a transform network ($f_w$) as a post processor for PCA models (Liu & Durlofsky, 2021):

$$\mathbf{m}_{\text{CNN-PCA}} = f_w(\mathbf{m}_{\text{PCA}}) \tag{5}$$

$\mathbf{m}_{\text{CNN-PCA}}$ is the resulting geological realization. For 3D models, a pre-trained C3D net (Tran et al., 2015) is applied as the transform network. The network is trained by minimizing the reconstruction error between $\mathbf{m}_{\text{PCA}}$ and $\tilde{\mathbf{m}}_{\text{PCA},i}$. Additional training losses such as a style loss and a hard data loss are considered as well. The detailed formulations of each loss component can be found in the original paper.

At the second step, we construct a single traditional PCA model for porosity and logarithmic permeability distributions. This practice is based on a common situation where porosity and permeability are correlated, but the specific correlation between them is unknown. For each realization of porosity and permeability distributions, $\mathbf{m} \in \mathbb{R}^{2N_c}$ is the concatenation of a normalized porosity vector and a normalized permeability vector, each with a length of $N_c$. The porosity and logarithmic permeability distributions are normalized based on their mean and standard deviation values in each facies of the $N_r$ realizations.

Fig.2 illustrates the two steps for generating a new reservoir model from a randomly sampled latent vector ($\lambda \in \mathbb{R}^{(n_1+n_2) \times 1}$). Here $n_1$ denotes the latent space dimension for the facies PCA model, and $n_2$ denotes the latent space dimension for the porosity/permeability PCA model. For the first step, a facies PCA realization is generated from the first $n_1$ components of latent vector $\lambda$, denoted as $\lambda_{n_1}$. A CNN-PCA facies realization is generated based on the trained C3D net. Since the CNN-PCA realization does not strictly follow the bimodal distribution, a histogram transformation is applied to fine tune the CNN-PCA realization to recover the same bimodal facies distribution followed by the initial $N_r$ model realizations. For the second step, we generate normalized porosity and permeability distributions from the last $n_2$ components of latent vector $\lambda$, denoted as $\lambda_{n_2}$. The final porosity and logarithm permeability distributions are obtained by rescaling the normalized porosity and permeability distributions per facies. In summary, the CNN-PCA method generates a facies distributions of different rock types and subsequently the traditional PCA method assigns porosity and permeability to the generated facies.

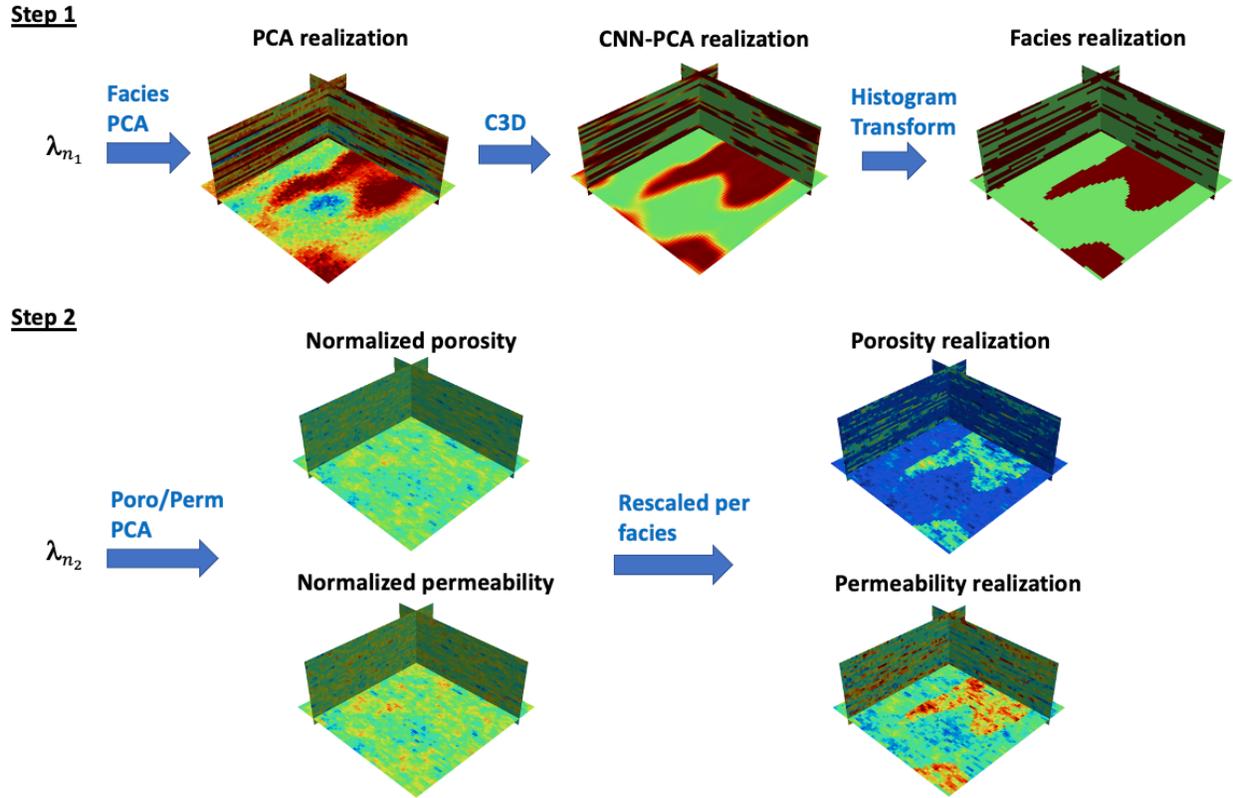

Fig.2. Illustration of the two steps of using CNN-PCA method to parameterize the reservoir model ensemble containing facies, porosity, and logarithmic permeability realizations.

2.2 Data parameterization

We use InSAR-inverted surface displacement map as our observation data. Each displacement map usually has a dimension of hundreds or thousands. Such a high-dimensional data space can lead to many problems such as extremely high computational cost and ensemble collapse issues in ensemble-based data assimilation methods. Therefore, we apply a similar traditional PCA approach as that presented in section 2.1 to reduce the dimension of the data space. To construct a PCA model, we form a data matrix $\mathbf{Y} \in \mathbb{R}^{N_y \times (N_p+1)}$ combining the observation data vector $\mathbf{y}_{\text{obs}}$ and predicted data vectors $\mathbf{y}_i$ ($i = 1, \ldots, N_p$). The predicted data vectors are the surface displacement predictions based on the prior reservoir model realizations as shown in Fig.1. $N_y$ denotes the data dimension, and $N_p$ denotes the number of the predicted data realizations.

$$\mathbf{Y} = \frac{1}{\sqrt{N_p}}\left[\mathbf{y}_{obs} - \boldsymbol{\mu}_y, \mathbf{y}_1 - \boldsymbol{\mu}_y, \dots, \mathbf{y}_{N_p} - \boldsymbol{\mu}_y\right], \qquad (6)$$

$\boldsymbol{\mu}_y$ is the mean of the $N_p$ +1 data vectors. A singular value decomposition yields $\mathbf{Y} = \mathbf{U\Sigma V^T}$. $\mathbf{U}$, $\mathbf{\Sigma}$ and $\mathbf{V}$ can be truncated accordingly to reduce the data space dimension from $N_y$ to $N_{\tilde{y}}$. The latent space observation data $\tilde{\mathbf{y}}_{obs} \in \mathbb{R}^{N_{\tilde{y}} \times 1}$ can be written as:

$$\tilde{\mathbf{y}}_{obs} = \mathbf{\Sigma}_{N_{\tilde{y}}}^{-1} \mathbf{U}_{N_{\tilde{y}}}^T (\mathbf{y}_{obs} - \boldsymbol{\mu}_y) = \mathbf{A}_{N_{\tilde{y}}}(\mathbf{y}_{obs} - \boldsymbol{\mu}_y). \qquad (7)$$

Considering the forward model to be $\mathbf{y}_{obs} = g(\mathbf{m}) + \mathbf{e}$, we can reformulate the problem in the latent data space as:

$$\tilde{\mathbf{y}}_{obs} + \mathbf{A}_{N_{\tilde{y}}}\boldsymbol{\mu}_{N_{\tilde{y}}} = \mathbf{A}_{N_{\tilde{y}}} g(\mathbf{m}) + \mathbf{A}_{N_{\tilde{y}}}\mathbf{e} \equiv \tilde{g}(\mathbf{m}) + \mathbf{e}_{\tilde{y}}, \qquad (8)$$

where $\mathbf{e}$ is the measurement error based on the original data vector and $\mathbf{e}_{\tilde{y}}$ is the measurement error based on the projected data vector. Since $\mathbf{A}_{N_{\tilde{y}}}$ is a linear transformation matrix, the projected covariance matrix of the observation data measurement error ($\tilde{\mathbf{C}}_y$) is correlated to the original error covariance matrix ($\mathbf{C}_y$) through (Grana et al., 2019):

$$\tilde{\mathbf{C}}_{\tilde{y}} = \mathbf{A}_{N_{\tilde{y}}} \mathbf{C}_y \mathbf{A}_{N_{\tilde{y}}}^T \qquad (9)$$

2.3 ES-MDA framework for model inversion

The ES-MDA approach updates model parameters (facies, porosity, and permeability) with observation data over several iterations to match the data to a satisfactory level. Here we introduce the implementation details of ES-MDA with an ensemble size of $N_e$.

- First, a prior ensemble of latent model parameters $\lambda^0 = [\lambda_1^0, \lambda_2^0, ..., \lambda_{N_e}^0] \in \mathbb{R}^{(n_1+n_2) \times N_e}$ is randomly sampled from the standard normal distribution. We can generate the corresponding PCA models and prior reservoir model ensemble as described in section 2.1.

- Second, we use a surrogate model (to be introduced in section 3.2) to predict surface displacement ($\mathbf{y}_i$) based on the prior reservoir model ensemble, where $i = 1, 2, ..., N_e$. We can obtain the projected data vector ($\tilde{\mathbf{y}}_i$ and $\tilde{\mathbf{y}}_{\text{obs}}$) and the projected covariance matrix of the data measurement error $\tilde{\mathbf{C}}_{\tilde{y}}$ following section 2.2.

- Third, the latent model parameter of the $i$th stochastic realization can be updated as follow:

$$\lambda_i^l = \lambda_i^{l-1} + \mathbf{C}_{\lambda\tilde{y}}^{l-1}\left(\mathbf{C}_{\tilde{y}\tilde{y}}^{l-1} + \alpha_{l-1}\tilde{\mathbf{C}}_{\tilde{y}}\right)^{-1}\left(\tilde{\mathbf{y}}_{uc,i}^{l-1} - \tilde{\mathbf{y}}_i^{l-1}\right), \tag{10}$$

where $l = 1, 2, ... N_a$, denoting the iterations. $\mathbf{C}_{\lambda\tilde{y}}^{l-1}$ is the cross-covariance matrix between the model parameters and the projected data given by:

$$\mathbf{C}_{\lambda\tilde{y}}^{l-1} = \frac{1}{N_e - 1}\sum_{i=1}^{N_e}(\lambda_i^{l-1} - \boldsymbol{\mu}_\lambda^{l-1})(\tilde{\mathbf{y}}_i^{l-1} - \boldsymbol{\mu}_{\tilde{y}}^{l-1})^T. \tag{11}$$

$\boldsymbol{\mu}_\lambda^{l-1}$ and $\boldsymbol{\mu}_{\tilde{y}}^{l-1}$ denote the mean of model parameter vector $\lambda_i^{l-1}$ and data vector $\tilde{\mathbf{y}}_i^{l-1}$.

$\mathbf{C}_{\tilde{y}\tilde{y}}^{l-1}$ is the auto-covariance matrix of the projected data given by:

$$\mathbf{C}_{\tilde{y}\tilde{y}}^{l-1} = \frac{1}{N_e - 1}\sum_{i=1}^{N_e}(\mathbf{y}_i^{l-1} - \boldsymbol{\mu}_{\tilde{y}}^{l-1})(\mathbf{y}_j^{l-1} - \boldsymbol{\mu}_{\tilde{y}}^{l-1})^T. \tag{12}$$

In Eq. (5), $\tilde{\mathbf{y}}_{uc,i}^{l-1}$ is the perturbed observation data (in the projected space) given by:

$$\tilde{\mathbf{y}}_{uc,i}^{l-1} = \tilde{\mathbf{y}}_{\text{obs}} + \sqrt{\alpha_{l-1}}\tilde{\mathbf{C}}_{\tilde{y}}^{1/2}\boldsymbol{\xi}_d^{l-1}, \tag{13}$$

where $\xi_d^{l-1} \sim N(0, \mathbf{I}_{N_d})$. $\alpha_{l-1}$ is the inflation coefficient, which satisfies the condition $\sum_{l=1}^{N_a} \frac{1}{\alpha_{l-1}} = 1$. In our application, we set the $\alpha$ of each iteration to $\frac{1}{N_a}$.

## 3. Numerical Forward Model and Surrogate Models

The workflow requires generating responses, namely displacement and pressure distribution, from any given reservoir realizations. This can be done using either a high-fidelity numerical simulator or a neural network-based surrogate model. The latter is usually trained using data generated by the former. In the following sections, we will introduce the numerical forward model and the detailed neural network structure for the surrogate models.

### 3.1 High-fidelity forward model

The high-fidelity forward model couples multi-phase flow in porous media and geomechanics to predict reservoir pressure and surface displacement due to $CO_2$ injection. The flow module describes the dynamics of $CO_2$ being injected in a deep saline aquifer. We consider two components ($CO_2$ and $H_2O$) transported by two phases (the aqueous phase and the gaseous phase). The gaseous phase only contains the $CO_2$ component and is in a supercritical state under subsurface conditions. The aqueous phase contains both $CO_2$ and $H_2O$ components. The generalized mass conservation equation of component $c$ is:

$$\phi \frac{\partial}{\partial t}\left(\sum_l \rho_l x_{cl} S_l\right) + \nabla \cdot \left(\sum_l \rho_l x_{cl} \mathbf{v}_l\right) - \sum_l \rho_l x_{cl} q_l = 0, \tag{14}$$

where $\phi$ is porosity, $l$ denotes different phases, $\rho_l$ is phase density, $S_l$ is phase saturation, $x_{cl}$ is the mass fraction of component $c$ in phase $l$ and $q_l$ is the injected phase volume. $\mathbf{v}_l$ is phase velocity, which can be calculated based on Darcy's law:

$$\mathbf{v}_l = -\frac{kk_{rl}}{\mu_l}(\nabla p_l - \rho_l g \nabla z), \tag{15}$$

where $k$ is the rock permeability, $k_{rl}$ is the phase's relative permeability, $\mu_l$ is phase viscosity, $p_l$ is phase pressure, $g$ is the gravitational acceleration constant, and $z$ is depth. The phase fractions and phase component fractions are calculated based on the model of Duan & Sun (2003). In the simulation, the phase densities and viscosities are updated via look-up in a precomputed table.

Surface displacement is calculated based on poromechanics (Biot, 1941). We solve the equilibrium equation of 3D porous media as follows:

$$\nabla(\boldsymbol{\sigma}' - b\tilde{p}\mathbf{I}) + \rho_b g \nabla z = 0, \tag{16}$$

where $\boldsymbol{\sigma}'$ denotes effective stress, $\mathbf{I}$ denotes the identity tensor, $b$ is the Biot coefficient, $\rho_b$ is the density of porous media, and $\tilde{p}$ is the equivalent pore pressure given by $\tilde{p} = \sum_l S_l p_l$. The effective stress tensor $\boldsymbol{\sigma}'$ is linked to the displacement field through a linear elastic constitutive relationship:

$$\boldsymbol{\sigma}' = \mathbf{C} : \nabla \mathbf{u} \tag{17}$$

$\mathbf{C}$ is the rank-four stiffness tensor, $\mathbf{u}$ is the solid displacement vector, and : is the inner product operator. In isotropic media, the stiffness tensor $\mathbf{C}$ (or $C_{ijkl}$) can be calculated as:

$$C_{ijkl} = K\delta_{ij}\delta_{kl} + G\left(\delta_{ik}\delta_{jl} + \delta_{il}\delta_{jk} - \frac{2}{3}\delta_{ij}\delta_{kl}\right), \tag{18}$$

where $\delta_{ij}$ is the Kronecker delta, $K$ is the bulk modulus and $G$ is the shear modulus. The pore pressure is first estimated by the flow module and then served as the input to the geomechanical module. The drained bulk modulus $K$ is correlated to reservoir porosity through pore compressibility ($c_\phi$):

$$K = \frac{1-\phi}{\phi c_\phi}. \tag{19}$$

In this work, an open source multiphysics simulator GEOSX is used (Settgast et al., 2018). Details of the numerical implementation and validation cases of the simulator can be found at http://www.geosx.org/. Although GEOSX can efficiently handle two-way full coupling, a one-way coupled approach is appropriate for our purpose: the pore pressure change and vertical effective stress change have a simple relationship as the vertical total stress remains largely constant.

3.2 Surrogate model based on 3D-to-2D R-U-Net

We develop two surrogate models in this section, one for surface displacement prediction and the other for depth-averaged reservoir pressure prediction. For both surrogate models, the output is a 2D map and the inputs are reservoir property distributions (3D fields) under a given injection scenario. We treat facies, porosity, and permeability fields in the storage reservoir as uncertain variables to be calibrated. Since the facies information is embedded in porosity and permeability distributions, we only use porosity and permeability distributions as inputs. Both surrogate models are based the same network structure as introduced below.

We propose a new 3D-to-2D network architecture inspired by a 3D-to-3D R-U-Net architecture applied in Tang et al. (2021b), where a U-Net is combined with several basic residual blocks. As shown in Fig.3(a), there are three input channels, each with a dimension of ($n_z$, $n_y$, $n_x$). The first channel is a normalized permeability field in a logarithmic scale, and the second channel is a normalized porosity field. The third channel is a 3D field with min-max normalized time values assigned to the grid cells along the well trajectories, while the values for the rest of the grid are zero. The encoding network extracts feature maps from input channels through several convolutional blocks. The encoded feature maps are then passed to four residual blocks. The feature map output from the last residual block is reduced by a mean operation in the depth direction and then fed into the decoding network. The reduced 2D feature maps are up-sampled through deconvolutional blocks in the decoding network to obtain a final surface displacement map. To improve prediction performance, the 3D feature maps in the encoding net are first reduced in the depth direction by the mean operation and then concatenated to the up-sampled 2D feature maps in the decoding net. These concatenation layers are key features of the U-Net architecture. The detailed network structure of the convolutional, residual and deconvolutional blocks are shown in Fig.3(b). In the deconvolutional blocks, we use up-sampling and convolutional layers instead of deconvolutional layers to minimize artifacts (Odena et al., 2016). The parameters for convolutional layers and up-sampling layers in each block are summarized in Table 1.

(a) 3D-to-2D R-U-Net architecture

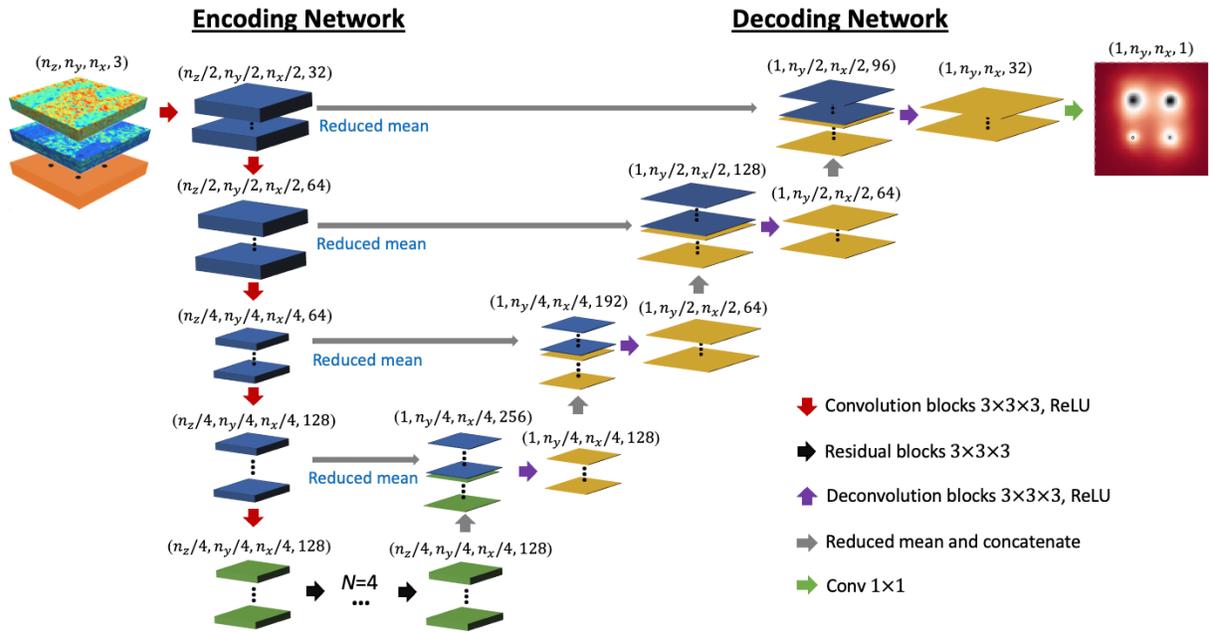

(b) Detailed structure of different blocks

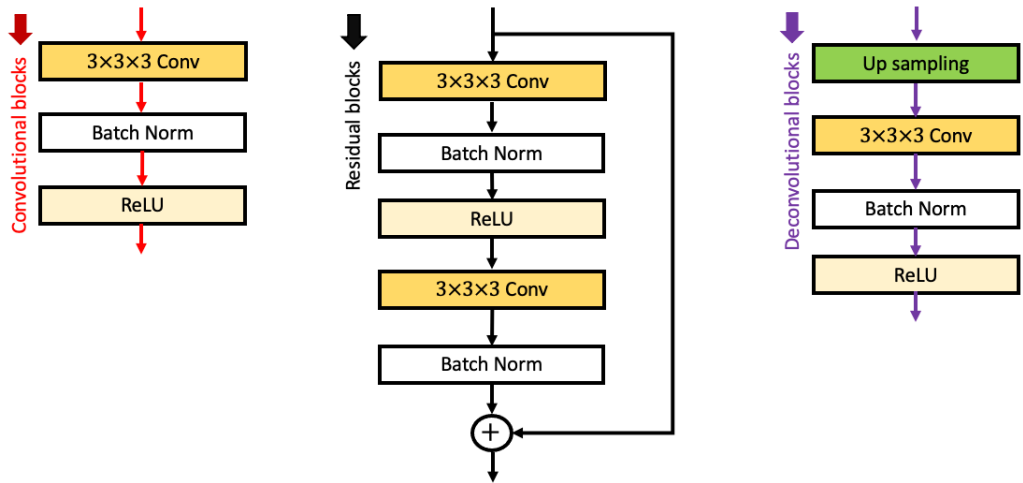

Fig.3. (a) Illustration of the 3D-to-2D R-U-Net architecture for surface displacement prediction. The input channels include a normalized logarithmic permeability map, a porosity map, and a map with pixel values at well locations being normalized time values and other pixel values being zero. $N=4$ indicates 4 residual blocks. (b) Illustration of the detailed network structure for convolutional, residual and deconvolutional blocks. Conv indicates a 3D convolution layer.

Table 1. Detailed architecture of 3D-to-2D R-U-Net.

| Name | Layer | Output size |
| --- | --- | --- |
| Encoder net | Input | $(n_z, n_y, n_x, 3)$ |

| | Conv, 32 filters of size 3 × 3 × 3, stride (2, 2, 2) | $(n_z/2, n_y/2, n_x/2, 32)$ |
|---|---|---|
| | Conv, 64 filters of size 3 × 3 × 3, stride (1, 1, 1) | $(n_z/2, n_y/2, n_x/2, 64)$ |
| | Conv, 64 filters of size 3 × 3 × 3, stride (2, 2, 2) | $(n_z/4, n_y/4, n_x/4, 64)$ |
| | Conv, 128 filters of size 3 × 3 × 3, stride (1, 1, 1) | $(n_z/4, n_y/4, n_x/4, 128)$ |
| Residual blocks | 128 filters of size 3 × 3 × 3, stride (1, 1, 1) | $(n_z/4, n_y/4, n_x/4, 128)$ |
| Decoder net | Up sampling, stride (1, 1, 1) <br> Conv, 128 filters of size 3 × 3 × 3 | $(1, n_y/4, n_x/4, 128)$ |
| | Up sampling, stride (1, 2, 2) <br> Conv, 64 filters of size 3 × 3 × 3 | $(1, n_y/2, n_x/2, 64)$ |
| | Up sampling, stride (1, 1, 1) <br> Conv, 64 filters of size 3 × 3 × 3 | $(1, n_y/2, n_x/2, 64)$ |
| | Up sampling, stride (1, 2, 2) <br> Conv, 32 filters of size 3 × 3 × 3 | $(1, n_y, n_x, 32)$ |
| | Conv, 1 filter of size 3 × 3 × 3, stride (1, 1, 1) | $(1, n_y, n_x, 1)$ |

a. $n_x$, $n_y$, $n_y$ indicate the dimension of the input 3D fields in x, y, and z directions respectively.
b. Conv indicates 3D convolution layer.

## 4. Application in a commercial-scale GCS model

The efficacy of the workflow is evaluated based on a synthetic commercial-scale GCS model. The model has four injection wells and a two million metric tons total injection rate per year. Model details and numerical results are introduced in the following sections.

4.1 Problem Setup

We build the reference reservoir model based on the clastic shelf depositional environment in Bosshart et al. (2018). Figs. 4(a) to 4(c) present the facies, porosity, and permeability distributions of the reference reservoir model. The reservoir model has a dimension of 32,156 × 32,156 × 85 m³, discretized into 64 × 64 × 28 cells. There are two facies in the reservoir, a shaly sand facies and a sand facies. Different facies honor different permeability curves as given in Bennion & Bachu (2007). Porosity and permeability values for each facies are sampled from the Energy & Environmental Research Center's (EERC) Average Global Database (AGD), which contains worldwide porosity/permeability measurements from deep saline aquifers. Fig.4(d) shows a cross plot between porosity and permeability based on each facies. Like most of the field cases, there is not a simple, one-to-one correlation between porosity and permeability.

The four injection wells (represented by the thick black lines in Fig.4(a)) has a total injection rate of 2 million metric tons of $CO_2$ per year. The four wells are grouped into a single manifold to honor the same tubing head pressure. Key reservoir properties are summarized in Table 2. The geomechanics model includes six overburden layers and one basement layer, of which the properties are summarized in Table 3. The reservoir is relatively thin compared with the vertical dimension of the formation. Note that all 3D visualizations of the reservoir in this paper are exaggerated by 100 times in the vertical dimension.

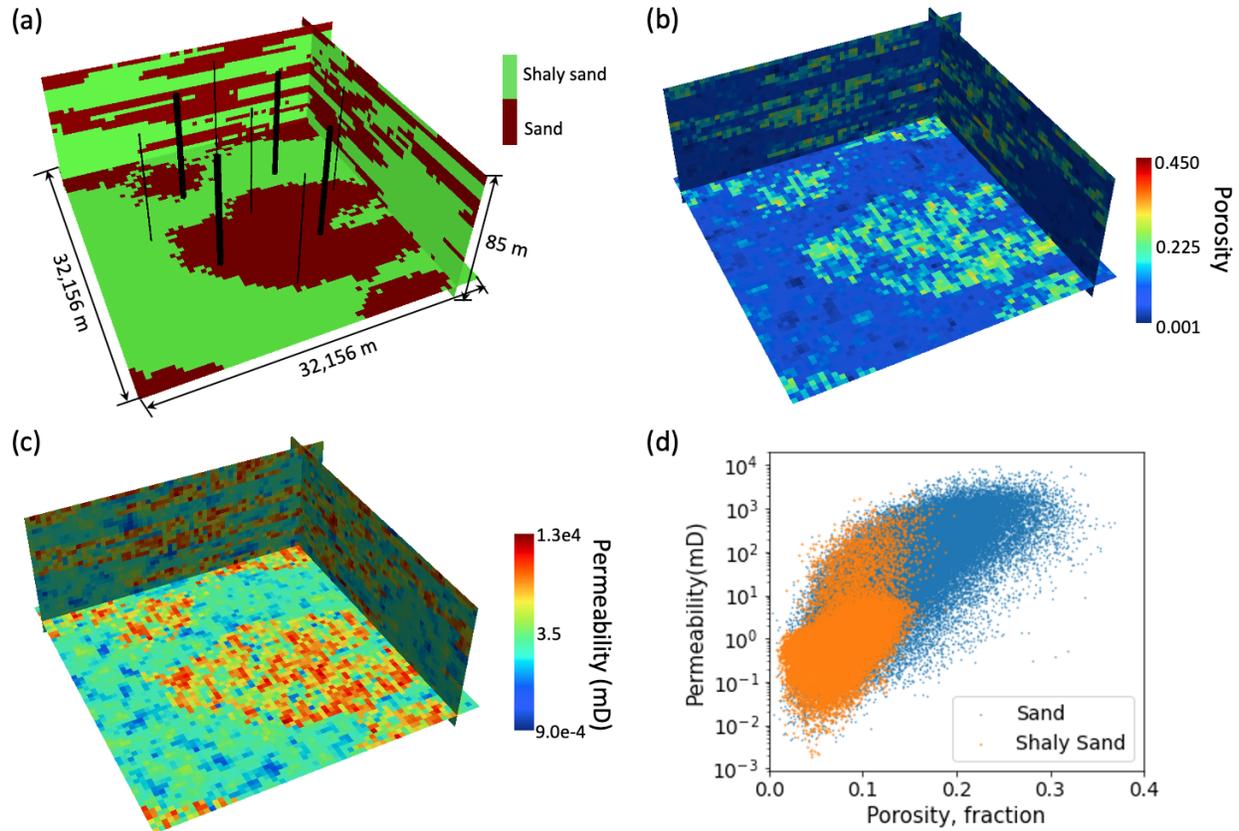

Fig.4. Reference reservoir model including (a) rock facies, (b) porosity and (c) permeability distributions. (d) A scatter plot between porosity and permeability. In Fig.4(a) the thick black line indicates injection wells, and the thin black line indicates exploration wells. The reservoir models are sliced at 5024 m in north and east directions from the boundaries and at 9 m from reservoir bottom.

Table 2. Reservoir properties used in GEOSX simulation

| Parameters | Value |
| --- | --- |
| Reservoir dimension | 32,156m × 32,156 m × 85 m |
| Grid resolution | 64 × 64 × 28 |
| Field injection target | 2 million metric tons per year |
| Pore compressibility | $4.64 \times 10^{-9}$ Pa$^{-1}$ |
| Relative permeability | High permeable sandstone and low perm clay/shale (Bennion & Bachu, 2007) |

Table 3. Formation properties for calculation of surface vertical displacement

| Top, m depth | Thickness, m | Rock Unit Description | Young's Modulus, GPa | Poisson's Ratio |
|---|---|---|---|---|
| 0 | 192 | Clays, silt, sand, gravel | 2.0 | 0.15 |
| 192 | 131 | Sandstone, mudstone | 10.0 | 0.20 |
| 323 | 625 | Shales | 16.2 | 0.23 |
| 948 | 53 | Sandstone | 10.0 | 0.20 |
| 1001 | 218 | Shales, mudstones, siltstones | 16.2 | 0.23 |
| 1219 | 85 | Storage reservoir | 9.9 – 18.9 | 0.20 |
| 1305 | 137 | Dolostone | 7.0 | 0.20 |

4.2 Geostatistical model and CNN-PCA realizations

To generate initial ensemble of reservoir models (i.e., informed guesses of reservoir property distributions), we use virtual measurements along nine well trajectories including four injectors (represented by the thick black lines in Fig.4(a)) and five exploration wells (represented by the thin black lines in Fig.4(a)). Virtual measurements are essentially queries of the reference reservoir model for rock facies, porosity, and permeability, which are referred as "hard data". We impose a 100% Gaussian noise in the measurement of porosity and permeability. Fig.5 shows the workflow to generate a single reservoir model realization from the hard data. Rock facies realizations are generated using Sequential Indicator Simulation (SIS). Normalized porosity and permeability realizations are generated using Sequential Gaussian Simulation (SGS) based on a selected random seed to preserve their dependency. Both SIS and SGS are conducted using the GSLIB software package (http://www.gslib.com/). The variogram parameters for SIS and SGS are summarized in Table 4. Both the variograms for SIS and SGS are assumed to be anisotropic with their major range and minor range sampled from a uniform distribution. After a rock facies realization is generated, porosity and permeability are rescaled based on the corresponding rock types.

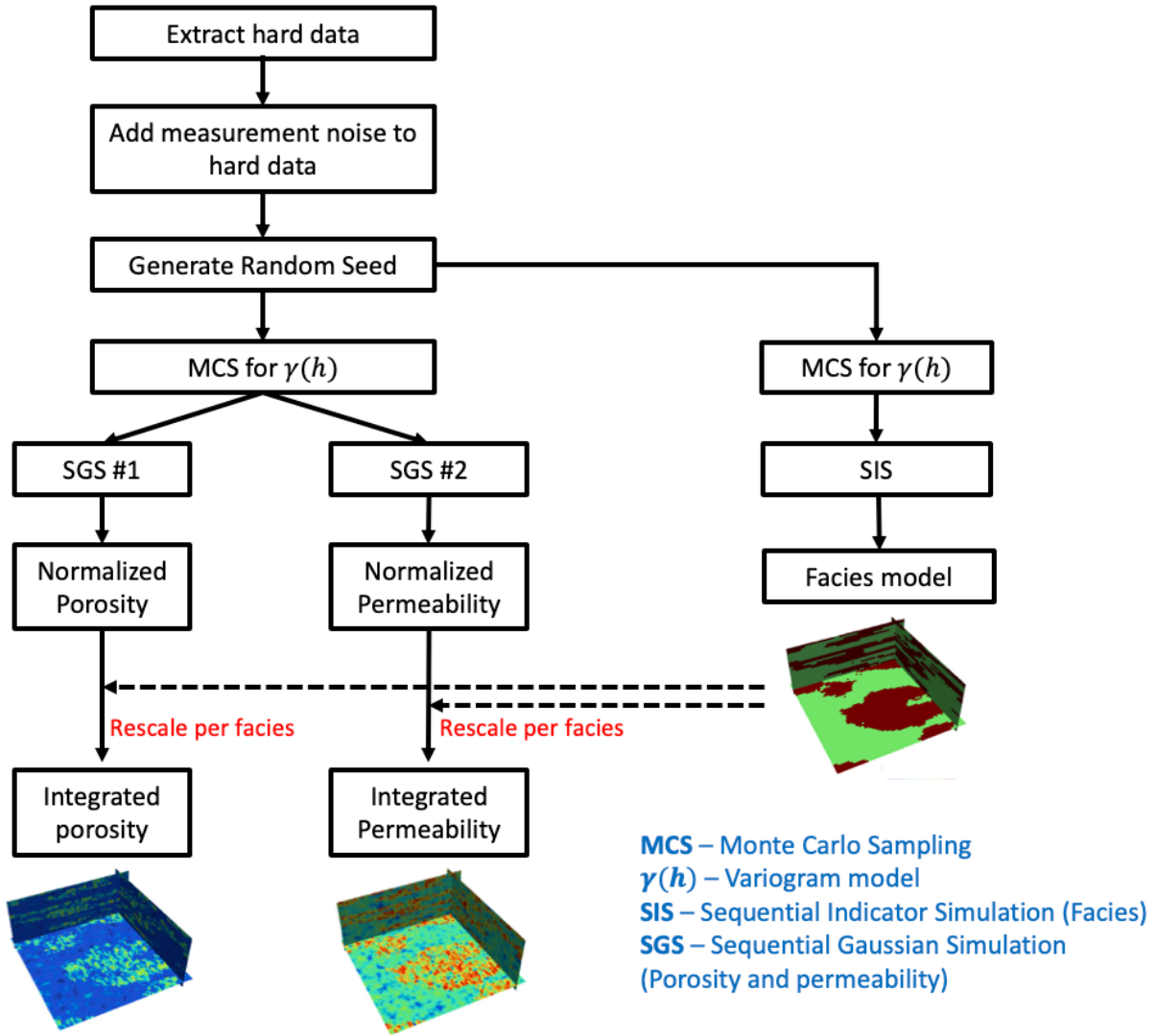

Fig.5. Workflow to generate a single reservoir model realization honoring hard data from the reference model.

Table 4. Variogram parameters for SIS and SGS.

| Model | Type | Major range, m | Minor range, m | Azimuth (major) |
|---|---|---|---|---|
| SIS | Exponential | $U(12561.0, 20097.6)$ | $U(5024.4, 7537.6)$ | 90° |
| SGS | Exponential | $U(1507.3, 2512.2)$ | $U(502.4, 1507.3)$ | 0° |

We generate 1000 initial reservoir models from the workflow and present three realizations in Fig.6(a). We conduct PCA model parameterization as introduced in section 2.1 based on these 1000 initial reservoir models. The reduced dimension for facies ($n_1$) is 290 with

70% of total variance being preserved ($T=70\%$). The reduced dimension for porosity and permeability ($n_2$) is 656 with 90% of total variance being preserved ($T = 90\%$). The total dimension of reduced parameter space is therefore 946. We reconstruct the reservoir model realizations in Fig.6(a) as shown in Fig.6(b).

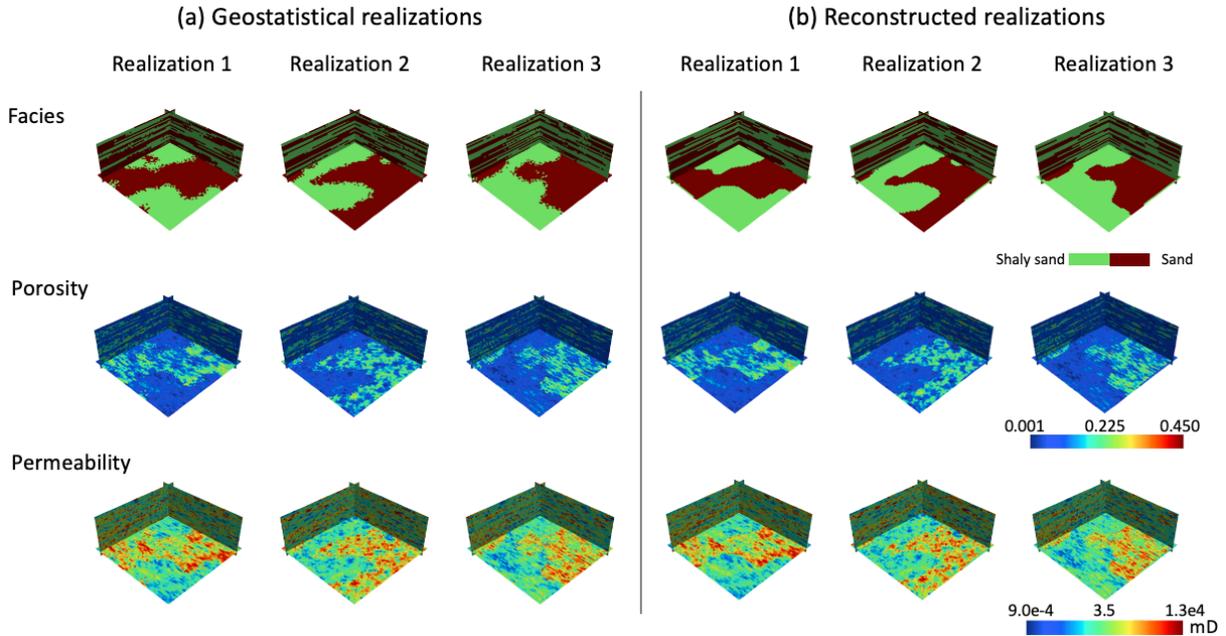

Fig.6. (a) Some geostatistical realizations of initial reservoir models and (b) the reconstructed realizations. A single reservoir model realization contains facies, porosity, and permeability distributions.

4.3 Training details for the surrogate models and model evaluation

We generate training, validation and testing datasets based on randomly sampled PCA realizations. The pressure maps are obtained by flow simulation recorded annually in a 10-year injection period. The surface displacement maps are obtained by geomechanical simulations with reservoir pressure imposed as boundary conditions after 1, 2, 5 and 10 years of injection. The training process is to minimize the training loss ($L_{train}$) including a reconstruction loss and a weighted loss. The weighted loss is calculated at well locations. The mismatch between numerical model predictions ($\mathbf{y}_i$) and neural network predictions ($\hat{\mathbf{y}}_i$) is calculated using an $L_2$ norm:

$$L_{\text{train}} = \frac{1}{n_m} \frac{1}{n_t} \sum_{i=1}^{n_m} \sum_{t=1}^{n_t} \left( \|\hat{\mathbf{y}}_i^t - \mathbf{y}_i^t\|_2^2 + \lambda \frac{1}{N_{wells}} \sum_{j=1}^{N_{wells}} \|\hat{\mathbf{y}}_{i,j}^t - \mathbf{y}_{i,j}^t\|_2^2 \right), \tag{20}$$

where $n_m$ is the number of training reservoir model realizations, and $n_t$ is the number of time steps ($n_t$ = 4 for displacement model; $n_t$ = 10 for pressure model). $\mathbf{y}$ is the surface vertical displacement for the displacement model and the depth-averaged reservoir pressure for the pressure model. We introduce extra weights to the grid cells containing wells to improve the accuracy for the pressure and displacement predictions. $N_{wells}$ is the number of wells and $\lambda$ is the weighting factor to emphasize prediction quality at the well locations. We apply $\lambda = 50$ for the displacement model and $\lambda = 100$ for the pressure model. The ADAM optimization algorithm with a learning rate of 0.001 and a batch size of 32 is applied. We choose the best model within 100 training epochs as the final model. The best model is determined as the model with the smallest loss based on the validation dataset. The networks are trained on a Nvidia Tesla P100 GPU for 320 min and 810 min for displacement and pressure models respectively for a training dataset consisting of 6,200 realizations.

We apply the root mean square error (RMSE) and the coefficient of determination ($R^2$) as the metrics to evaluate model accuracy:

$$\text{RMSE} = \sqrt{\frac{1}{n_s} \frac{1}{n_t} \sum_{i=1}^{n_s} \sum_{t=1}^{n_t} \|\hat{\mathbf{y}}_i^t - \mathbf{y}_i^t\|_2^2} \tag{21}$$

$$R^2 = 1 - \frac{\sum_{i=1}^{n_s}\sum_{t=1}^{n_t}\|\hat{\boldsymbol{y}}_i^t - \boldsymbol{y}_i^t\|_2^2}{\sum_{i=1}^{n_s}\sum_{t=1}^{n_t}\|\hat{\boldsymbol{y}}_i^t - \overline{\boldsymbol{y}}_i^t\|_2^2} \tag{22}$$

where $n_s = 100$ is the number of testing realizations. We test the accuracy of surrogate models trained with different numbers of training realizations, $n_m$=1,000, 2,800, 4,500 and 6,200. Fig.7 presents the RMSE and $R^2$ metrics for displacement and pressure models. As $n_m$ increases from 1000 to 6,200, the displacement model accuracy increases from an $R^2$ value of 0.956 to 0.985, and pressure model accuracy increases from an $R^2$ value of 0.943 to 0.983. We apply the displacement model and the pressure model trained by 6,200 realizations in the following analyses and the workflow.

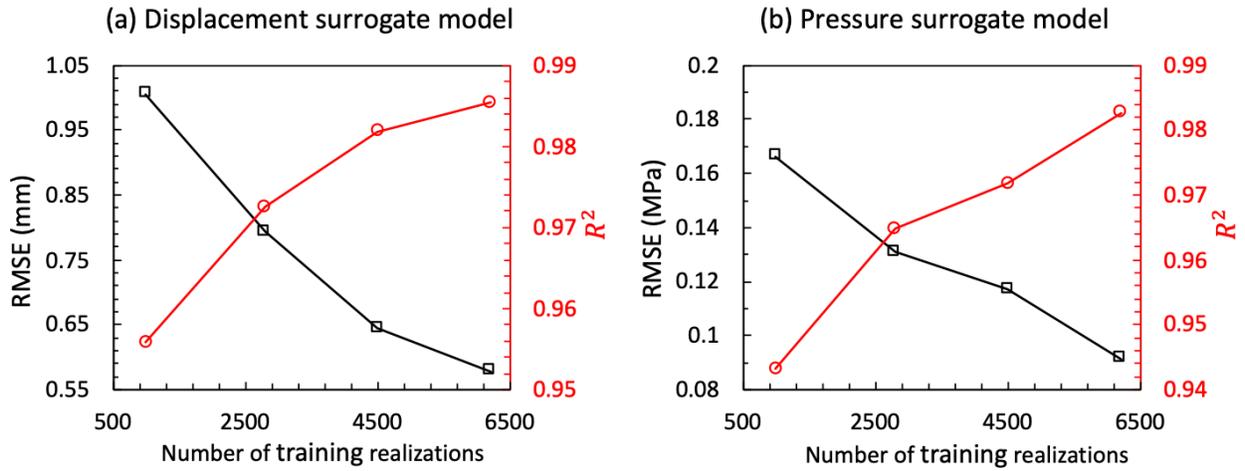

Fig.7. Performance metrics of (a) displacement surrogate model and (b) pressure surrogate model.

We further demonstrate the surrogate models' performance for predicting reservoir responses to 10 years of injection based on five randomly selected realizations in the testing dataset as shown in Fig.8. We arrange the results in an ascending order of $R^2$ based on the displacement model. The $R^2$ values for the displacement model are 0.885, 0.949, 0.962, 0.981 and 0.986, and the $R^2$ values for the pressure model are 0.895, 0.959, 0.980, 0.984 and 0.994,

respectively. Overall, the surrogate models have satisfactory predictive performance over a wide range of reservoir scenarios. The underperformance of realization 1 is mainly because the scenario where the upper right well takes almost no injection is not sufficiently represented in the training dataset. This scenario happens when the permeabilities along a wellbore are poorly connected vertically, which therefore reduces the injectivity of the well. We can also observe that the surface displacement distribution is highly correlated to the depth-averaged pressure distribution in each realization.

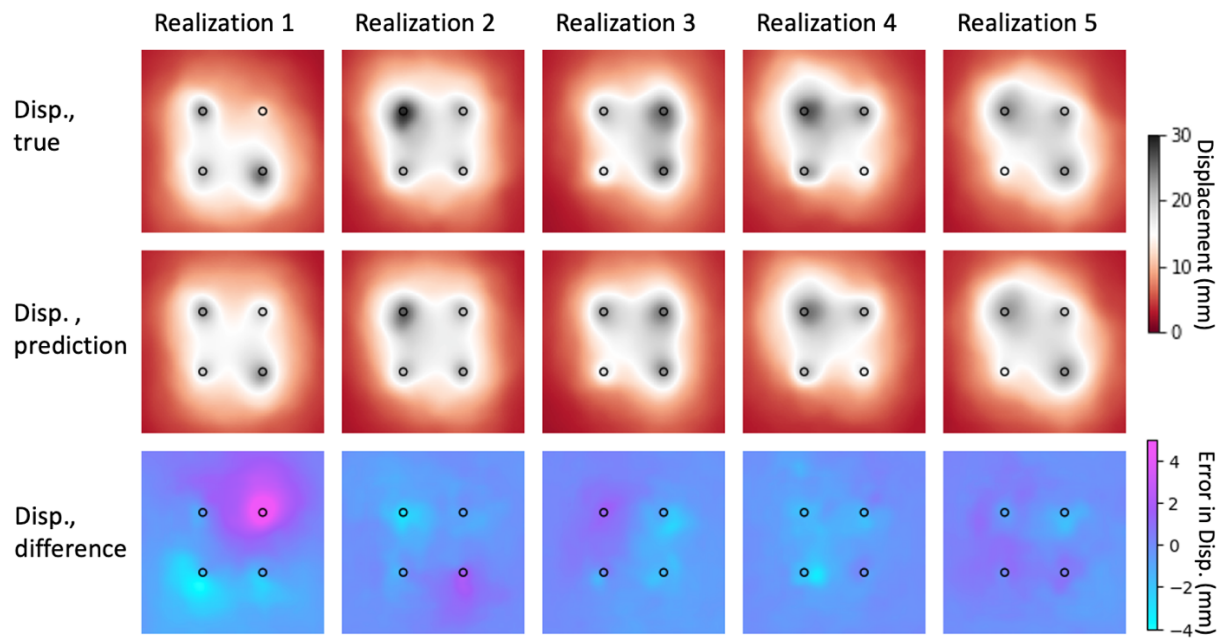

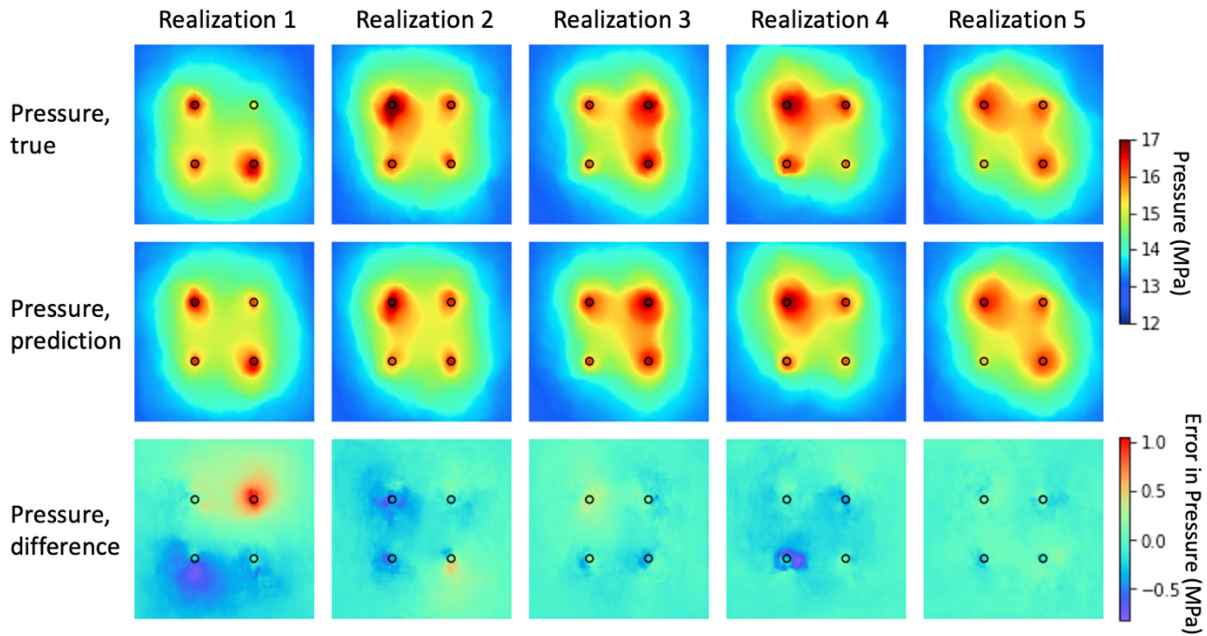

Fig.8. (a) Surface displacement predicted by simulation (true) and the displacement surrogate model (prediction) after ten years of injection. (b) Depth-averaged reservoir pressure predicted by simulation (true) and the pressure surrogate model (prediction) after ten years of injection. Difference = prediction – true.

Fig.9 shows the model performance in the time domain based on a single, randomly selected realization. The injection durations being evaluated are 2, 6, 10, 12, and 16 years. The $R^2$ values for the displacement model are 0.941, 0.947, 0.955, 0.955 and 0.937 respectively, and the $R^2$ values for the pressure model are 0.930, 0.942, 0.942, 0.946 and 0.956 respectively. Recall that the displacement model is trained using simulation results after 1, 2, 5 and 10 years of injection. The results illustrate that the surrogate models have accurate interpolation capability within the 10-year injection period. Both surrogate models have acceptable extrapolation performance for up to 16 years of injection.

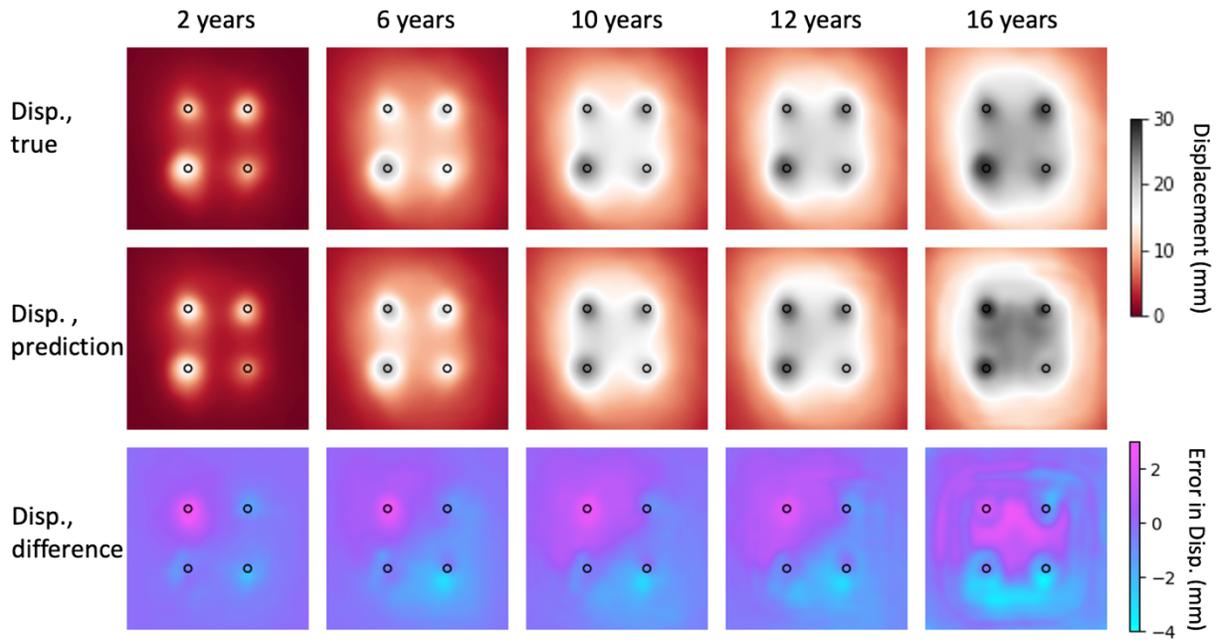

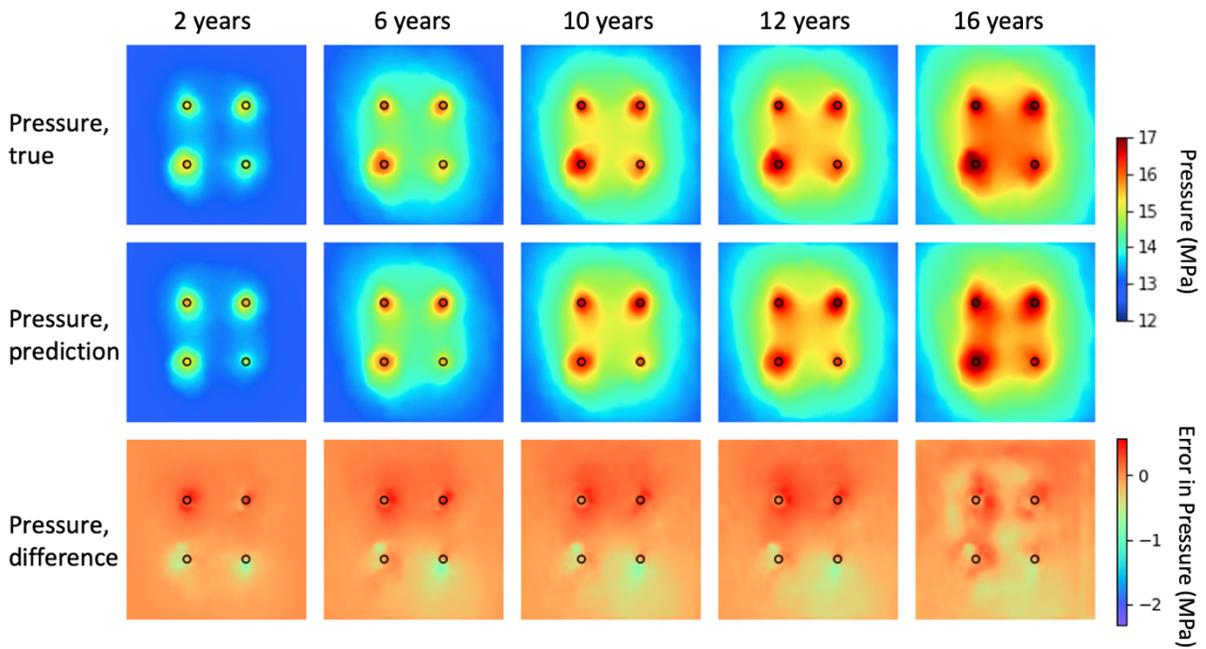

Fig.9. (a) Surface displacement predicted by simulation (true) and the displacement surrogate model (prediction) during 16 years of injection. (b) Depth-averaged reservoir pressure predicted by simulation (true) and the pressure surrogate model (prediction). Difference = prediction – true.

4.4 Data assimilation and forecasting results

We follow the workflow shown in Fig.1 to perform data assimilation and forecasting. We use an ensemble size $N_e = 100$ and an iteration number of $N_a = 12$ for ES-MDA. As mentioned in section 4.2, the parameter space dimension of this problem is 946 based on CNN-PCA parameterization. We consider surface vertical displacement at the ground surface after 2 years of injection as the observation data. We will use the observation data to calibrate the reservoir models to provide reservoir performance predictions at the observation time (2 years) and future times (after 2 years). A Gaussian white noise with a standard deviation of 5% of the observation values is applied to the simulated data to form the observation data as shown in Fig.10. We apply the data parameterization method introduced in section 2.2 to reduce the data space dimension. In this application, $N_p = 100$ and $N_y = 4096$. With 90% of the total variance being preserved ($T$ = 90%), the data space can be reduced to $N_{\tilde{y}} = 11$ latent variables for each realization.

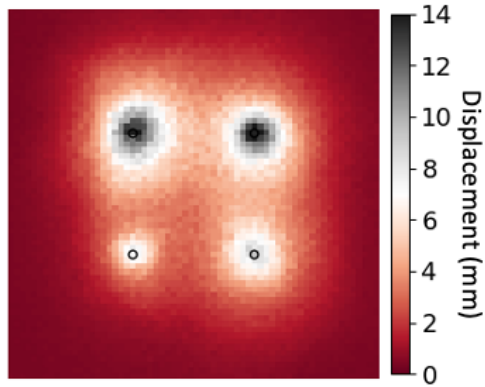

Fig.10. Observation data: surface vertical displacement after two years of injection.

Fig.11 presents the prior and posterior reservoir model ensembles' surface displacement predictions after two and ten years of injection. We use RMSE to quantify the history matching prediction performance. In this application, $\hat{\mathbf{y}}_i$ denotes the mean predicted surface displacement map base on the posterior ensemble and $\mathbf{y}_i$ denotes the "true" surface displacement map. The

RMSE values are 0.27 and 0.86 mm for the two prediction periods. The results demonstrate that the posterior reservoir model ensemble can provide accurate surface displacement predictions at both the observation and future timesteps with significantly reduced uncertainties.

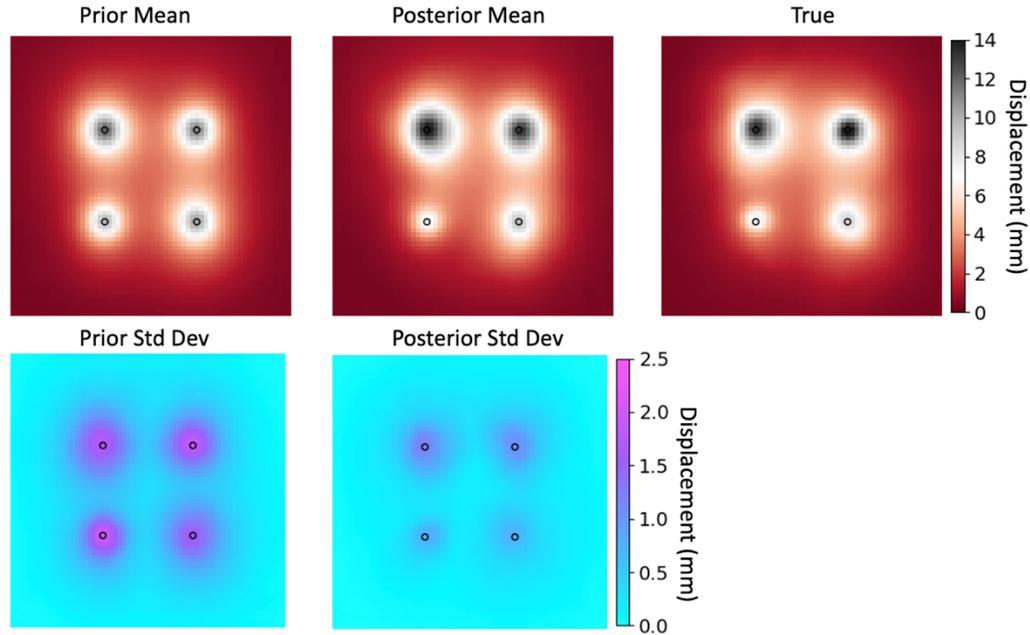

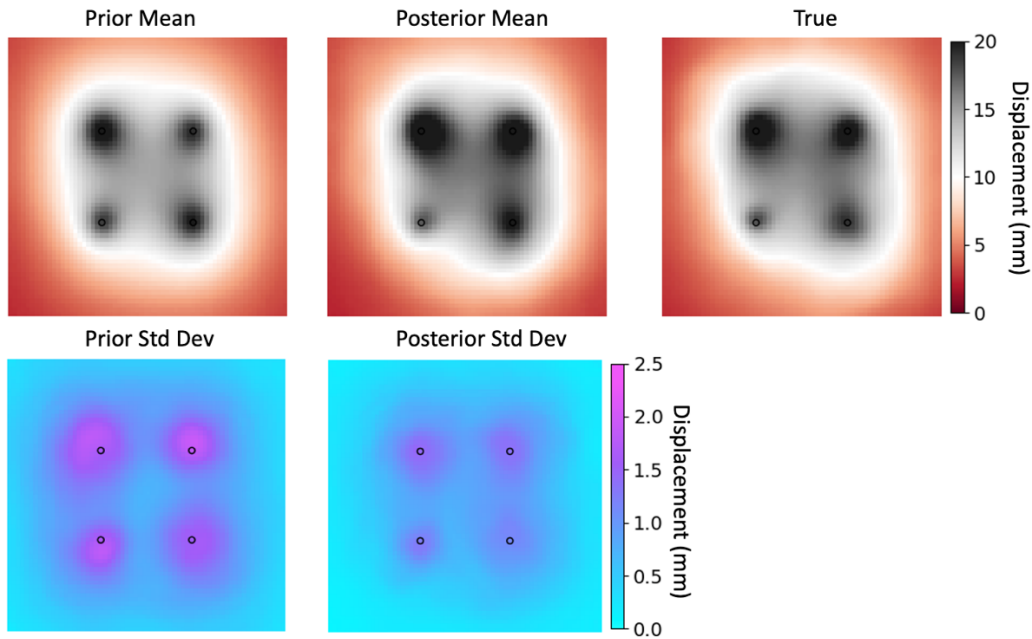

Fig.11. Surface displacement predictions after (a) two years and (b) ten years of injection. Prior and posterior predictions are compared with true surface displacement map from numerical

simulation. The mean and standard deviation maps are calculated based on the entire prior or posterior reservoir model ensemble.

To further examine whether the workflow can provide reasonable reservoir pressure predictions, we apply the pressure surrogate model to predict depth-averaged pressure maps after 2, 5 and 10 years of injection based on the posterior reservoir model assemble. As shown in Fig.12, the mean and standard deviation of each prediction ensemble is compared with the "true" pressure map from numerical simulation. We use RMSE to quantify the prediction performance. In this application, $\hat{\mathbf{y}}_i$ is the mean predicted pressure map from the posterior ensemble and $\mathbf{y}_i$ is the "true" pressure map. The RMSE values are 0.08, 0.11, and 0.15 MPa respectively for three prediction periods. The results illustrate that the posterior reservoir models calibrated by surface displacement observation can provide reasonable predictions of reservoir pressure dynamics at current and future injection periods.

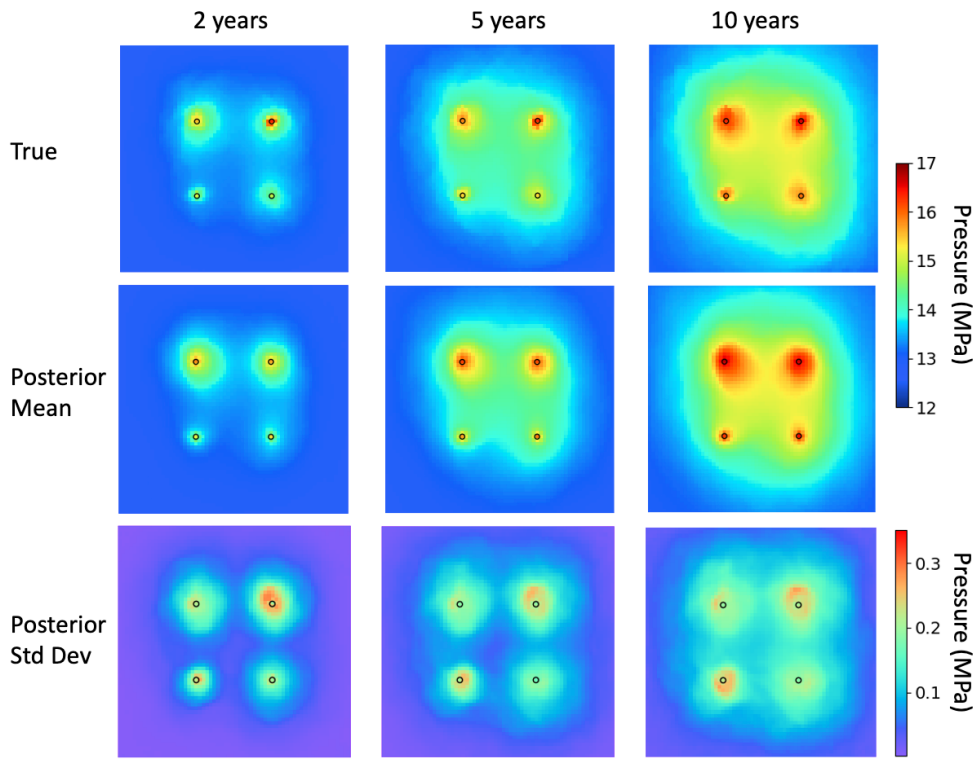

Fig.12. Depth-averaged reservoir pressure predictions after 2, 5 and 10 years of injection. The true pressure map is from numerical simulation. The mean and standard deviation maps are calculated based on the entire posterior reservoir model ensemble.

Fig.13 compares prior and posterior reservoir models in three realizations. A wide variety of facies distributions are considered in the prior ensemble. After data assimilation, the facies distributions are constrained to a narrower range. Fig.14 presents the mean and standard deviation maps of the prior porosity / permeability ensembles and the posterior porosity / permeability ensembles. The reference porosity and permeability maps are provided for comparison. After history matching, the posterior reservoir models share more common features with the reference model and the uncertainties are significantly reduced. Note that exactly matching the reference model is not the goal of our workflow. First, due to the high dimensionality of a reservoir model, an inversion problem is always ill-posed. Second, in a real-world history matching problem, the ground truth reservoir model is unknown. The key metric for success is the ability of the constrained posterior models to accurately predict reservoir pressure distribution in response to injection.

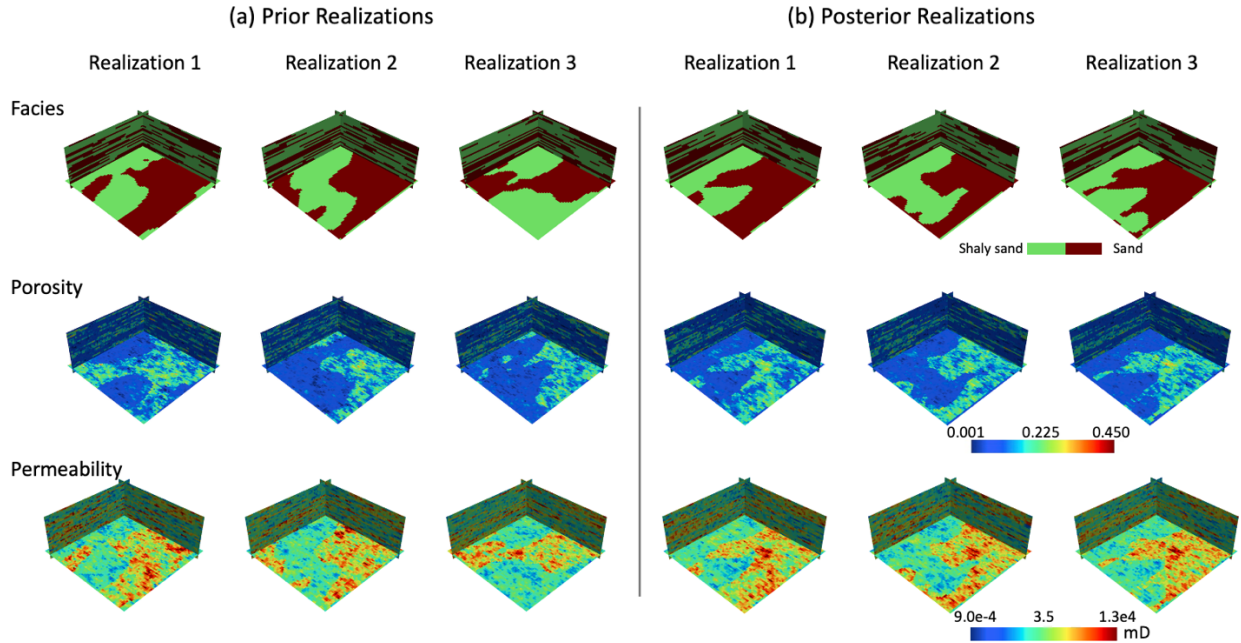

Fig.13. Three randomly selected realizations of (a) prior and (b) the corresponding posterior reservoir models. Each reservoir model realization contains facies, porosity, and logarithmic permeability distributions.

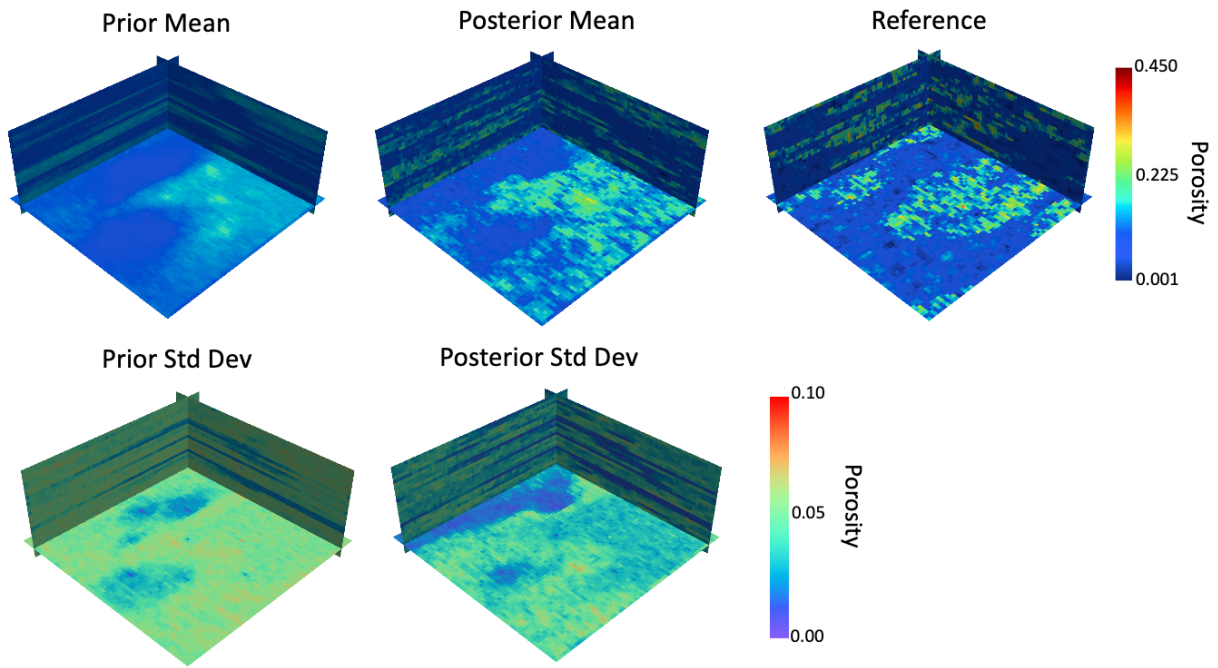

(a) Prior and posterior porosity distributions

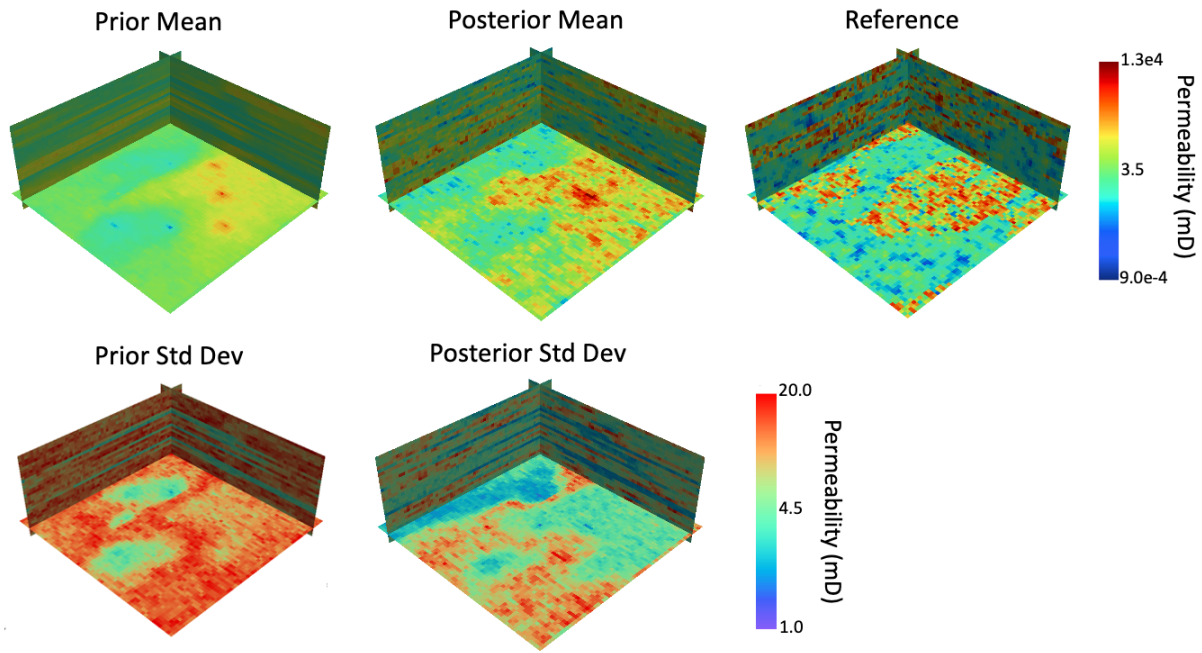

Fig.14. Mean and standard deviation of prior / posterior (a) porosity and (b) permeability ensembles. Reference porosity and permeability fields are shown for comparison.

4.5 Effect of observation data noise level and reduced data space dimension

In this section, we analyze the effect of two factors on the performance of data assimilation and forecasting: observation data noise level and the degree of data space dimension reduction. As for the observation data noise level, we evaluate two Gaussian white noise levels: 7% and 10% of the observation values in addition to the baseline 5%. The other parameters of the workflow are kept the same as the previous analysis. We calculate the $R^2$ values of displacement and pressure predictions of each realization in the ensemble compared with the relevant true maps. Table 5 presents the 95% confidence interval of the calculated $R^2$ for injection durations of 2 and 10 years. The results show that increasing the Gaussian white noise level to 10% of the observation values will not significantly compromise the prediction performance. A higher noise level can lead to a wider uncertainty range of predictions.

Table 5. The 95% confidence interval of $R^2$ values for surface displacement and pressure predictions after 2 and 10 years of injections, evaluated at different Gaussian white noise levels.

| Noise level | $R^2$ values of 2yr displacement | $R^2$ values of 10yr displacement | $R^2$ values of 2yr pressure | $R^2$ values of 10yr pressure |
|---|---|---|---|---|
| 5% | [0.967, 0.972] | [0.936, 0.951] | [0.913, 0.924] | [0.923, 0.937] |
| 7% | [0.955, 0.962] | [0.919, 0.938] | [0.896, 0.908] | [0.907, 0.926] |
| 10% | [0.924, 0.946] | [0.880, 0.919] | [0.899, 0.913] | [0.916, 0.938] |

As for reduced data space dimension, we consider two additional preserved PCA variance percentages $T = 94\%$ and $T = 97\%$, in addition to $T = 90\%$. With $N_p = 100$ being fixed, $T = 94\%$ results in a data space dimension of 15, and $T = 97\%$ results in a data space dimension of 24. A trial-and-error process reveals that we must increase the ES-MDA ensemble size ($N_e$) to avoid ensemble collapse issues as the data space becomes larger. We set $N_e = 1000$ for $T = 94\%$ and $N_e = 3000$ for $T = 97\%$. For different preserved variance percentages, we summarize the 95% confidence interval of $R^2$ values for displacement and pressure predictions after 2 and 10 years of injection in Table 6. For both $T = 94\%$ and $T = 97\%$, the prediction performance is slightly better than that of $T = 90\%$. Preserving an appropriate percentage of variance is important: if the preserved variance percentage is too low, some key features of surface displacement maps might be missed, thereby undermining the fidelity of the inversion results. On the other hand, if the preserved variance percentage is excessively high, some local features (such as the measurement noises) might be preserved to interfere with the inversion algorithm.

Table 6. The 95% confidence interval of $R^2$ values for surface displacement and pressure predictions after 2 and 10 years of injections, evaluated at different preserved variance percentages ($T$) for data space dimension reduction.

| $T$ | $R^2$ values of 2yr displacement | $R^2$ values of 10yr displacement | $R^2$ values of 2yr pressure | $R^2$ values of 10yr pressure |
|---|---|---|---|---|
| 90% | [0.967, 0.972] | [0.936, 0.951] | [0.913, 0.924] | [0.923, 0.937] |
| 94% | [0.976, 0.977] | [0.963, 0.965] | [0.911, 0.914] | [0.956, 0.957] |
| 97% | [0.973, 0.974] | [0.941, 0.942] | [0.933, 0.934] | [0.950, 0.952] |

4.5 Computational cost

Table 7 summarizes the computational and training cost of the data assimilation and forecasting workflow. With an ensemble size of 100 and an iteration number of 12, the workflow can take approximately 3,000 core-hours to finish with GEOSX, a high-fidelity reservoir model, as the forward simulator. If we replace the forward simulator with the displacement and pressure surrogate models, the workflow runs in approximately half an hour on a 4-core laptop. The computational cost has been transferred to the generation of training datasets for the surrogate models, which costs approximately 14,000 core-hours. In a field application, the training process can be performed before data become available. The significantly accelerated data assimilation workflow allows multiple trials to determine the appropriate parameters for the workflow, which can significantly improve its efficiency in field applications.

Table 7. Summary of the computational and training cost of the data assimilation workflow

| Time | Physical Simulator | DL Based Surrogate Models |
|---|---|---|
| Single run (pressure) | ~2 core-hours[a] | 0.06 core-s[b] |
| Single run (displacement) | ~0.25 core-hours[a] | 0.06 core-s[b] |
| Data assimilation (100 ensemble, 12 iteration) | ~2,925 core-hours[a] | ~0.5 machine-hours[b] |
| Training dataset cost | - | ~13,950 core-hours[a] |
| Training | - | 15.5 GPU-hours / 5.4 GPU-hours[c] (pressure / displacement) |

a. On an Intel Xeon E5-2695 v4, single-core serial run.
b. On MacBook Pro, i5-8279U CPU, utilizing up to four cores.
c. On an NVIDIA Tesla P100 GPU.

## 5. Concluding Remarks

In this study, we develop a computationally efficient workflow to forecast reservoir pressure in 3D carbon storage reservoirs by assimilating surface displacement data. The surface displacement data is a 2D map that can be obtained from low-cost InSAR measurements. We employed an ensemble-based framework, ES-MDA, to drive the data assimilation. Various PCA techniques are applied to reduce the dimensionality of the parameter and data spaces. The significantly reduced parameter and data spaces boost the computational speed of the workflow by reducing the ensemble size required by ES-MDA. The workflow is also accelerated by two deep learning-based surrogate models for surface displacement and reservoir pressure predictions. The surrogate models are demonstrated to have good accuracy and generalizability to different durations of $CO_2$ injection.

The workflow is evaluated against a synthetic commercial-scale GCS model with two facies. We apply a two-step PCA approach to tackle the parameterization challenge introduced by heterogeneous distributed permeability and porosity in individual facies. The same approach is also applicable to reservoir models with more than two facies. The workflow relies on additional well logs to provide hard data for the geostatistical model. We noticed for the synthetic case study, logs from the four injection wells alone are not sufficient to properly constrain the statistical models. Adding five virtual exploration wells successfully addressed this issue. This is not surprising as the concerned domain in the model is quite large, and a larger reservoir naturally requires more information to constrain. In practice, other data sources, such as a baseline 3D seismic inversion could be used to further constrain the models.

The efficacy of current workflow depends upon reliable surface displacement maps obtained from InSAR data inversion. A limitation of the current workflow is that the displacement maps we used were synthetic. Although a moderate amount of white noise was superposed, they perhaps are still "cleaner" than real InSAR data. The real-world application may involve additional challenges. For example, in a region with significant surface vegetation or other significant ground changes, the InSAR data between sweeps becomes decorrelated, leading to blocks of missing data. Correlated noise is also common due to some seasonal variations such as near-surface groundwater recharge. These are important "data engineering" issues to be addressed before the proposed workflow can be effectively applied to real-world applications. The contribution of the current work is mainly a "proof of concept"; a more robust solution accommodating the aforementioned issues would require more research and development effort in the future.

## Acknowledgements

This manuscript has been authored by Lawrence Livermore National Security, LLC under Contract No. DE-AC52-07NA2 7344 with the U.S. Department of Energy (DOE). The United States Government retains, and the publisher, by accepting the article for publication, acknowledges that the United States Government retains a non-exclusive, paid-up, irrevocable, world-wide license to publish or reproduce the published form of this manuscript, or allow others to do so, for United States Government purposes. This report is LLNL-JRNL-822252. This work was completed as part of the Science-informed Machine learning to Accelerate Real Time decision making for Carbon Storage (SMART-CS) Initiative (edx.netl.doe.gov/SMART). Support for this initiative came from the U.S. DOE Office of Fossil Energy's Carbon Storage Research program. Funding for GEOSX development was provided by TotalEnergies. through the FC-MAELSTROM project and the U.S. Department of Energy, Office of Science, Exa-scale Computing Project. FH completed this work during a visiting scientist appointment at LLNL.